# Robust Optimal Task Planning to Maximize Battery Life

Jiachen Li*, Chu Jian*, Feiyang Zhao*, Shihao Li*, Wei Li*, Dongmei Chen*

* *The University of Texas at Austin, TX 78721 USA (email: jiachenli, jian_chu, feiyang_zhao, shihaoli01301@utexas.edu, weiwli@austin.utexas.edu, dmchen@me.utexas.edu)*

**Abstract**: This paper proposes a control-oriented optimization platform for autonomous mobile robots (AMRs), focusing on extending battery life while ensuring task completion. The requirement of fast AMR task planning while maintaining minimum battery state of charge, thus maximizing the battery life, renders a bilinear optimization problem. McCormick envelop technique is proposed to linearize the bilinear term. A novel planning algorithm with relaxed constraints is also developed to handle parameter uncertainties robustly with high efficiency ensured. Simulation results are provided to demonstrate the utility of the proposed methods in reducing battery degradation while satisfying task completion requirements. [1]

## 1. INTRODUCTION

In various industry and commercial settings, autonomous mobile robots (AMR) have been prevalent for automating tasks such as material handling and transportation. These AMRs primarily use batteries as their power source. Ensuring efficient battery operations and extending battery life while meeting overall system performance requirements is crucial for reducing costs and maintenance expenses. Extensive research has been conducted on extending battery life, particularly in the areas of battery modeling [Doyle et al. (1993); Fuller et al. (1994)], charging and discharging strategy exploration, degradation quantification, and control development. Empirical approaches, as demonstrated in [Kim and Ha (1997), Moss et al. (2008)], use experimental data to establish relationships between battery degradation and certain factors, such as temperature and charging and discharging rates.

Optimizing charging transfer dynamics, including charging and discharging strategies, plays a crucial role in prolonging battery life [Grolleau et al. (2014), Lam and Bauer (2013)]. Capacity fading, referred to as cycling loss, is one of the major degradation mechanisms related to charging and discharging strategies. Cycling losses, as discussed in [Choi and Lim (2002)], indicate that fast charging, which typically involves high charging current rates, accelerates battery degradation. An optimal tradeoff can be achieved between minimizing the charging rate and prolonging the battery life. Large depth of discharge [Ramadass et al. (2003)] and high temperature [Tredeau and Salameh (2009)] are also factors that accelerate battery degradation and have been intensively studied. For degradation quantification, the degradation rate is typically modeled with quadratic [Ning and Popov (2004)], or linear [Spotnitz (2003)] functions with these affecting factors.

It should be noted that battery capacity fading, referred to as calendar loss, also occurs when the battery is stored at elevated state of charge (SOC) without charging or discharging dynamics [Rong and Pedram (2006)]. Research has demonstrated that calendar losses are a function of SOC [Kim and Ha (1997)]. When in operation, an AMR travels between stops or remains idle to perform certain activities. Capacity fading due to calendar losses, which occur during idle time, should be considered. Maintaining a low SOC at idle will minimize battery degradation.

Therefore, for an AMR that traverses and performs certain tasks at different stops, it is ideal to have adequate SOC to complete the next tasks without overcharging the battery.

Additionally, most of the studies on minimizing battery degradation have focused on deterministic problem settings [Hoke et al. (2011)], where the required battery power or energy is known. However, in practical applications, battery usage requirements are often uncertain. To address these uncertainties, two approaches have been developed. The first one uses a drive cycle as a benchmark, similar to the automotive industry, to ensure optimal performance for a given task [Kivekas et al. (2018)]. Nevertheless, the robustness of this approach depends on the generality of the driving cycle. The second approach is using robust optimization, which optimizes the worst-case scenario across various environments. In [Korolko and Sahinoglu (2017)], robust optimization techniques are presented for optimal charging schedules. However, the uncertainty of the required battery power is not discussed.

To optimize the AMR performance and battery life extension under the uncertainties of AMR power and SOC requirement, a comprehensive optimization framework is proposed to 1) minimize battery degradation during idle while ensuring task completion; and 2) determine the battery operating constraints based on their characteristics to seek robust solutions against uncertainties while maintaining high efficiency.

To achieve the first objective of minimizing battery degradation while ensuring task completion, a bilinear term that includes the charging rate, charging time, and battery idle time is used in the objective function. This bilinear term arises from the interaction between charging rates and time, whose product is a function of the required SOC. Both elements are integral to the objective function. This innovative approach to extending battery longevity represents a novel formulation of optimal control problem.

---

[1] This is the default footer for the provided file.

Furthermore, to achieve the second objective of seeking robust optimization against uncertainties, a two-stage optimization structure is needed. The first stage is dedicated to optimizing the charging rate, AMR speed, and target SOC, all without prior information regarding the tasks. The second stage refines these optimizations once task uncertainties are clarified. The goal is to determine the optimal output from the first stage, thus achieving the best possible results under conditions of uncertainty. Unlike traditional robust control strategies, this two-stage optimization framework tolerates a degree of constraint violation, offering a more efficient approach under real-world conditions.

In this research, tractable algorithms have been developed for solving a nonlinear, optimal, bilinear control problem with a bilinear term in real-time. Here "tractable" refers to the ability to solve the problem using standard optimization solvers. Most optimal control/optimization problems in similar settings are usually NP-hard and do not have an analytical solution. In these situations, Genetic Algorithm (GA) and Neural Network are often applied to obtain numerical solutions, which can be time-consuming and result in local optimal solutions. To make the proposed problem tractable with a low computational cost, different linearization and reformulation techniques are explored. As a result, this work delivers a computationally efficient optimization framework that enables systematic battery energy management while ensuring optimal power utilization and extended battery life under uncertain operating conditions.

## 2. PROBLEM FORMULATION

The objective of this research is to develop an optimization framework to minimize battery cycling degradation and idling capacity loss while completing the tasks an AMR is assigned.

### 2.1 Battery Modeling

Two battery degradation mechanisms are modeled in this work to characterize different aspects of battery aging. In addition to the cycling degradation, the calendar aging or idling capacity loss is also captured in the battery model.

The cycling degradation of the battery is characterized by the Arrhenius equation [Lam and Bauer (2013)], which outlines the relationship between battery degradation and selected stress factors, especially the battery discharge rate that is defined as C-rate.

$$\eta_{cyc,i} = \left(k_1 SOC_{dev,i} \cdot e^{k_2 \cdot SOC_{avg,i}} + k_3 e^{k_4 \cdot SOC_{dev,i}}\right) \cdot Q_i \quad (1)$$

where $\eta_{cyc,i}, Q_i$ represent the degradation of battery and the capacity of the battery in one charging cycle. The parameters $k_i$ are estimated from experimental data. Simulation results indicate that, within the same charging cycle, the impact of C rate on battery degradation is significantly greater than that of SOC. Consequently, it is reasonable to disregard the degradation effect of SOC during the charging period.

The calendar aging of the battery is described by the capacity fade model, which is a function of SOC and time $t$ [Grolleau et al. (2014)].

$$\frac{dQ_{\text{loss}}}{dt} = k(T, SOC) \cdot \left(1 + \frac{Q_{\text{loss}}(t)}{C_{\text{nom}}}\right)^{-\alpha(T)} \quad (2a)$$

where $k(T, SOC)$ is given by:

$$k_A \cdot e^{-\frac{E_A}{R}\left(\frac{1}{T} - \frac{1}{T_{\text{ref}}}\right)} \cdot SOC + k_B \cdot e^{-\frac{E_B}{R}\left(\frac{1}{T} - \frac{1}{T_{\text{ref}}}\right)} \quad (2b)$$

$k_A, k_B, E_A, E_B$ are experimentally determined parameters, while $R, T_{\text{ref}}$ represent the gas constant and reference temperature, respectively, and can be considered as constants.

By varying the C-rate and SOC, different degradation phenomena can be simulated. With this empirical model, and assuming constant temperature $T$, the coefficients of battery degradation can be derived as a function of the C-rate and SOC ($\eta_{\text{charging}} = f_c(c), \eta_{\text{idle}} = f_s(SOC)$).

### 2.2. AMR application

The system operates within a workspace containing fixed task locations where both task execution and battery charging can occur. The AMR has a workflow as shown in Fig. 1. After being assigned different groups of tasks, the AMR executes tasks, recharges itself to the target SOC $\bar{S}$, and waits for the next group of tasks to come. $t_i, t_{c_i}, t_{w_i}, \Delta t_i$ represent the execution time for task group $i$ (consider moving from the task point i to task point i+1 is also contain in the excuting time), charging time after task group $i$, AMR idle time after task group $i$, and postponed time for task group $i$, respectively. Note that, $\Delta t_1 = 0$ since task 1 is executed immediately. $d_i$ represents the estimated distance for task group $i$. $\xi_i$ represents the time interval between two task group $i$ and $i + 1$.

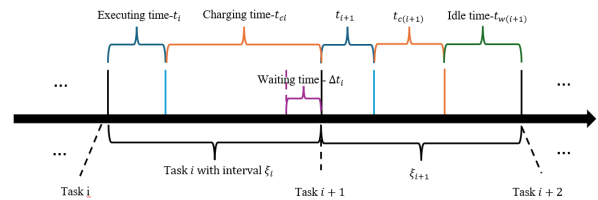

Figure 1: Defining AMR variables

In the baseline/benchmark scenario, the target SOC is set to $\bar{S} = 0.8$, the AMR speed $v$ is set to $\bar{v}$, and the C-rate $c$ is set to $\bar{c}$. This configuration ensures that tasks and battery recharging are completed as fast as possible.

### 2.3. Objective Function

Based on the model, an objective function is formulated in this section. The objective is to simultaneously minimize battery degradation and ensure task completion. Achieving this objective involves determining an appropriate low target SOC, selecting an optimal C-rate, and a quick transition to new tasks.

Consider a set with $N$ groups of tasks, each group $i$ is assigned with a time window $\xi_i$. For each task group, two different aspects of degradation are accounted for as discussed in 2.1. The first one is associated with battery charging (3), and the second one is from battery staying idle (4), for $i = 1, 2, ..., N$

$$\eta_{c_i} = f_c(c) \cdot t_{c_i} \quad (3)$$
$$\eta_{s_i} = f_s(\bar{S}) \cdot t_{w_i} \quad (4)$$

where $t_{c_i}, t_{w_i}$ denotes charging and idle time in task group $i$. For completing tasks, the task waiting time $\sum \Delta t_i$ needs to be minimized. Therefore, the overall objective function is:

$$\min_{\bar{S}, v, c} \sum_{i=1}^{N} (\eta_{c_i} + \eta_{s_i} + \lambda \Delta t_i) \quad (5)$$

By minimizing the objective function, the decision variables $\bar{S}, v, c$ (also defined as control inputs) are determined to give the optimal target SOC, robot velocity and C-rate. Compared to the baseline setup, using these control inputs protects the battery in the long run because the optimization considers the aging associated with battery being at elevated SOC during idle.

*2.4. Constraints*

Followed by the problem formulation, several constraints are considered. For simplicity, capacity consumption is assumed to be proportional to the velocity $v$ as

$$\Delta SOC = k_v \cdot v \cdot t_i \quad (6)$$

where $k_v$ is the coefficient of capacity consuming rate.

The battery capacity has a lower bound $\underline{S}$ that guarantees the battery to have the capability to handle some unpredictive circumstances. Therefore, the consumable battery capacity for executing the tasks should be lower than $\bar{S} - \underline{S}$ as

$$\Delta SOC \leq \bar{S} - \underline{S} \quad (7)$$

After executing a group of tasks, the battery needs to be recharged, therefore the charging capacity is equal to the capacity consumed as

$$\Delta SOC = c \cdot t_{c_i} \quad (8)$$

The task waiting time and idle time will be calculated recursively by

$$\Delta t_{i+1} = \max\{0, \Delta t_i + t_i + t_{c_i} - \xi_i\} \quad (9)$$
$$t_{w_{i+1}} = \max\{0, \xi_i - t_i - t_{c_i} - \Delta t_i\} \quad (10)$$

Let $T$ be the decision variables related to time $(t, t_c, t_w, \Delta t)$, the constraints is obtained in a compact matrix form as:

$$K_T \cdot T \geq [\xi, 0] \quad (11)$$

The physical limits are also considered, such as AMR speed limit and C-rate limit.

$$v \in [v^L, v^U], c \in [c^L, c^U] \quad (12)$$

The task executing time and charging time are bounded as

$$t_i \in [t^L, t^U], t_{c_i} \in [t_c^L, t_c^U] \quad (13)$$

*2.5. Model Linearization*

A model has been developed to describe a battery powered AMR incorporating several nonlinear characteristics. The presence of bilinear terms in both the objective function and the constraints renders the optimization problem NP-hard. To address this highly challenging NP-hard problem, various linearization methods can be employed to approximate the nonlinear terms effectively.

Compared to the task of executing time $t_i$, robot speed has a narrower range, therefore, linearizing $v$ is performed. Take the first order Taylor Expansion to $\frac{1}{v}$ around a nominal value $\hat{v}$, the linearized approximation is obtained as:

$$t_i = d_i \cdot \left(-\frac{1}{\hat{v}^2} \cdot v + \frac{2}{\hat{v}}\right) \quad (14)$$

Similarly, constraint (8) is linearized as:

$$t_{c_i} = k_v \cdot d_i \cdot \left(-\frac{1}{\hat{c}^2} \cdot c + \frac{2}{\hat{c}}\right) \quad (15)$$

The battery degradation coefficients $f_c(c), f_s(\bar{S})$ are functions of C-rate $c$ and target SOC $\bar{S}$, an approximated linear relation can be derived from simulation results of (1-2), therefore, the degradation calculation is simplified as:

$$\eta_c = k_c \cdot c \cdot t_{c_i} \quad (16)$$
$$\eta_s = k_s \cdot \bar{S} \cdot t_{w_i} \quad (17)$$

In (17), where the product of $\bar{S}$ and $t_{w_i}$ presents a bilinear term without any known parameters to directly facilitate its simplification, it can be approximated using the piecewise McCormick Envelope technique [Castro (2015)], yielding equation:

$$\begin{cases} w_i \geq \bar{S} \cdot t_k^L + S_j^L \cdot t_{w_i} - S_j^L \cdot t_k^L - M \cdot (1 - z_{il}) \\ w_i \geq \bar{S} \cdot t_k^U + S_j^U \cdot t_{w_i} - S_j^U \cdot t_k^U - M \cdot (1 - z_{il}) \\ w_i \leq \bar{S} \cdot t_k^L + S_j^L \cdot t_{w_i} - S_j^U \cdot t_k^L + M \cdot (1 - z_{il}) \\ w_i \leq \bar{S} \cdot t_k^U + S_j^U \cdot t_{w_i} - S_j^L \cdot t_k^U + M \cdot (1 - z_{il}) \end{cases} \quad (18)$$

where $w_i = \bar{S} \cdot t_{w_i}$, $\sum_{l=1}^{N_S \cdot N_t} z_{il} = 1$, interval bounds are given by $S_j^L = S^L + (S^U - S^L) \cdot \frac{j-1}{N_S}$, $S_j^U = S^L + (S^U - S^L) \cdot \frac{j}{N_S}$, $t_k^L = t_w^L + (t_w^U - t_w^L) \cdot \frac{k-1}{N_t}$ and $t_k^U = t_w^L + (t_w^U - t_w^L) \cdot \frac{k}{N_t}$, where $S^L$, $S^U$, $t_w^L$ and $t_k^U$ are the overall bounds, $j = 1, 2, ..., N_S$ and $k = 1, 2, ..., N_t$.

Equation (18) is rewritten into a compact matrix form ($e$ denotes vector or matrix with all ones) as:

$$w - \bar{S} \cdot T - S \cdot t_w + ST + M \circ (e - z) \geq 0 \quad (19a)$$
$$w - \bar{S} \cdot T - S \cdot t_w + ST - M \circ (e - z) \leq 0 \quad (19b)$$

*2.6. Summary of the Bilinear Optimization*

The complete Mixed-Integer Linear Programming (MILP) formulation consists of the following components: the

objective function maintains its original form of minimizing battery degradation and task completion time:

$$\min_{\bar{S},v,c} \sum_{i=1}^{N} (\eta_{c_i} + \eta_{s_i} + \lambda \Delta t_i) \qquad (20)$$
$$s.t. (7), (11-15), (19)$$

The optimal control inputs, $\bar{S}, v, c$, will be obtained as results of the bilinear optimization.

### 3. PROBLEM WITH UNCERTAINTY

The above problem formulation is for deterministic systems. In reality, the AMRs operate with uncertainties. This section introduces modeling of uncertainties in required battery power and discusses the application of a chance-constrained reformulation to address these uncertainties.

#### 3.1. Model uncertainty

The uncertainty in required battery power can be represented by the uncertainty of task interval $\xi$ and required distance $d$, let $\delta \in R^{2N}$ denotes the generalized uncertainty, therefore, $\Delta \xi_i = e_i^T \delta, \Delta d_i = e_{N+i}^T \delta$, where $e_i$ denotes the basis vector with all entries being 0 expect for the $i$-th entry equals to 1. Furthermore, defining $e_\xi^T \delta = \Delta \xi$ and $e_d^T \delta = \Delta d$, where $e_\xi \triangleq \{e_1, e_2, \dots, e_N\}, e_d \triangleq \{e_{N+1}, e_{N+2}, \dots, e_{2N}\}$. The vector of cost coefficients for adjustments is denoted by $\beta$. To effectively incorporate uncertainty into (20), a two-stage robust optimization is proposed by making a "here and now" decision, which will make the decision at the first stage, without prior knowledge of the uncertainty $\delta$. Then, once $\delta$ is observed, a second stage action will be adapted based on the observed $\delta$. Here, a linear decision rule (LDR) [Kuhn et al. (2009)] is adopted. It approximates second-stage decisions as linear functions of observed uncertainties. The essence of this problem is as follows: the $1^{st}$-stage control inputs $(\bar{S}, v, c)$ must be determined in the absence of any information about the uncertainty. The $2^{nd}$ − stage control inputs are decided upon once the uncertainty becomes known. The framework shown below demonstrates how information flows through the optimization process.

The problem under uncertainty can be formulated as

**Algorithm 1** Two-Stage Robust Optimization
1: **Input:** Task data, battery parameters, uncertainty distribution
2: **Output:** Optimal first-stage decisions, adaptive second-stage decisions
3: **First Stage: Here-and-Now Decisions**
4: Generate $K$ uncertainty samples $\delta^k$
5: Solve SAA problem:
6:    Minimize SAA objective (degradation, task time, violation penalties)
7:    Subject to deterministic and sample-specific constraints
8:    With chance constraint: $\sum_{k=1}^{K} g^k \leq \epsilon K$
9: Obtain $\bar{S}^*, v^*, c^*$
10: **Second Stage: Adaptive Decisions**
11: Observe realized uncertainty $\delta$
12: Adjust using LDR: $S_{\text{adj}} = \bar{S}^* + A\delta$, $v_{\text{adj}} = v^* + B\delta$, $c_{\text{adj}} = c^* + C\delta$
13: Deploy adjusted decisions

The control variables of the second stage can be viewed as responses to the uncertainty $\delta$. The reformulated problem (21a) aims to minimize the battery degradation and task waiting time at the first stage, which is the same as (5), as well as the expected extra cost for the second stage actions.

$$\min_{\bar{S},v,c} \sum_{i=1}^{N} (\eta_{c_i} + \eta_{s_i} + \lambda \Delta t_i) + E[\beta^T \delta] \qquad (21a)$$
$$s.t. (7), (11-15), (19)$$
$$t^L \leq t + W_t \delta \leq t^U \qquad (21b)$$
$$t_c^L \leq t_c + W_{tc} \delta \leq t_c^U \qquad (21c)$$
$$0 \leq k_v(d + e_d^T \delta) \leq \bar{S} + W_s \delta - \underline{S} \qquad (21d)$$
$$K_T(T + W_T \delta) \geq [\xi + e_\xi^T \delta, 0] \qquad (21e)$$
$$w - (\bar{S} + W_s \delta)T - S(t_w + W_{tw}\delta)$$
$$+ ST + M \circ (e - z) \geq 0 \qquad (21f)$$
$$w - (\bar{S} + W_s \delta)T - S(t_w + W_{tw}\delta)$$
$$+ ST - M \circ (e - z) \leq 0 \qquad (21g)$$

#### 3.2. Chance-constrained Formulation

In practical AMR operations, not all constraints can or should be treated with equal rigidity. Some constraints, such as basic operational limits, must be strictly satisfied, while others, particularly those mainly affected by uncertainties, may allow for occasional violations of the constraints without significantly compromising system performance. This practical consideration leads to the adoption of two complementary approaches:

The chance-constrained formulation allows certain constraints to be satisfied probabilistically rather than deterministically. For instance, while the AMR must maintain sufficient battery charge for basic operations, it may occasionally operate below the ideal SOC target without system failure.

The constraints that can be violated are organized in the set $\mathcal{V} = \{(21b - 21e)\}$. Let $X$ denotes $1^{st}$ stage variables $(\bar{S}, v, c, t, t_c)$, and the set $\mathcal{V}$ can be simplified as

$$\mathcal{V} = \{A(X)\delta \leq B(X)\} \qquad (22)$$

Assuming the constraints imposed on the task-executing time, charging time, and SOC lower bound can be violated. Their chance constraints can be written as

$$\mathcal{P}\{A_v(X)^T \delta \leq B_v(X)\} \geq 1 - \epsilon, \forall v \in \mathcal{V} \qquad (23)$$

For the constraints such as $(21f - 21g)$, that cannot be violated, the dualization techniques in [Ben-Tal et al. (2004)] are used to reformulate them into:

$$TW_S + SW_{tw} - D_1^T U = 0 \qquad (24a)$$
$$w - \bar{S}T - St_w - ST + M \circ (e - z) \geq D_1^T t \qquad (24b)$$
$$TW_S + SW_{tw} + D_2^T U = 0 \qquad (24c)$$
$$w - \bar{S}T - St_w - ST - M \circ (e - z) \leq D_2^T t \qquad (24d)$$

where $D_{1,2}$ are the dual variables, $U, t$ denote the uncertainty constraint ($U\delta \leq t$). By dualization, the uncertainty $\delta$ is removed from the constraint, which makes the problem tractable.

#### 3.3. Sample Average Approximation (SAA)

However, chance constraints are generally difficult to evaluate directly, especially when the underlying uncertainty distribution is unknown or complex. This computational challenge motivates the use of Sample Average

Approximation (SAA), which approximates the probabilistic constraints using a finite set of uncertainty samples.

Consider existing $K$ samples $\{\delta^k\}$ with $k \in \{1,2,\ldots,K\}$, the SAA method will use these sample-based empirical distribution to replace the unknown distribution $\mathcal{P}$, which assigns equal probability to all samples. In other words, all the samples need to satisfy the probability constraint.

Therefore, under the empirical distribution, the problem introduces new mixed-integer constraints, and (23) will be equivalent to the following constraints.

$$g_v^k \in \{0,1\} \quad (25a)$$
$$A_v(X)^T \delta^k - B_v(X) - M \cdot g_v^k \geq 0 \quad (25b)$$
$$\sum_{k=1}^{K} g_v^k \leq K \cdot \epsilon, \forall v \in \mathcal{V} \quad (25c)$$

*3.4. Summary of Robust Optimization*

The robust optimization framework developed in Sections 3.1-3.3 transforms the deterministic MILP into a two-stage decision process that accounts for uncertainties in task intervals and required distances. The first stage determines the baseline control variables before uncertainties are revealed, while the second stage provides adaptive responses to observed uncertainties through linear decision rules.

The objective function incorporates both the deterministic cost and expected uncertainty impact:

$$\min_{\bar{S},\bar{v},\bar{c}} \sum_{i=1}^{N} \left(\eta_{c_i} + \eta_{s_i} + \lambda \Delta t_i\right) + E[\beta^T \delta]$$

$$s.t. (7), (11-15), (19), (24), (25)$$

The SAA method discretizes the uncertainty space using K samples, introducing binary variables $g_v^k$ to control constraint violations while maintaining the specified probability threshold $\epsilon$. The dualization approach eliminates uncertainty terms from strict constraints through the introduction of dual variables $(D_1, D_2)$ and uncertainty set parameters $(U, t)$.

## 4. NUMERICAL RESULTS AND DISCUSSION

In this section, numerical results are presented to validate the proposed framework. Results for different methods are provided and compared to that of the benchmark strategy. All the numerical simulations were conducted on a regular desktop equipped with Intel CPU i7 10700 K 3.8GHz and 16 GB of RAM using MATLAB 2022a and MOSEK 10.2.

*4.1. Optimization with Uncertainty*

Under the chance constraints, the optimization result for target SOC is presented in Fig. 2, which indicates a few violations of the SOC lower bound constraints. The violation rate is determined by the probability threshold $\epsilon$, which is chosen according to the specific operating scenario. When $\epsilon = 0.02$ is selected, the result for violation probability of each task is presented in Fig. 3. Most tasks have a violation probability well below this limit, indicating they satisfy the chance constraint, while only a few tasks approach or exceed the limit, as also observed in Fig. 3. Figure 4 shows the AMR's operational schedule under the optimized solution with uncertainty.

In comparison to the baseline strategy ($\bar{S} = 0.8, \bar{v}, \bar{c}$), the model with chance-constrained consideration achieves significantly lower degradation rates with relatively low $\bar{S}$ (~0.55) and $c$ as expected. Though the proposed approach introduces approximately 6 seconds more waiting time than the baseline, it delivers a 70.3% reduction in calendar degradation. Another observation is that the target SOC ($\bar{S}$) is higher when the uncertainty level increases.

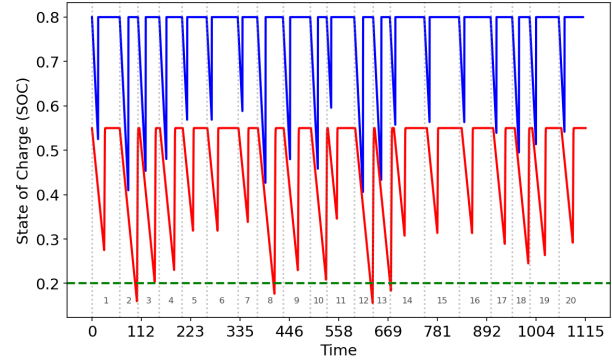

Figure 2: The target SOC in 20 tasks testing

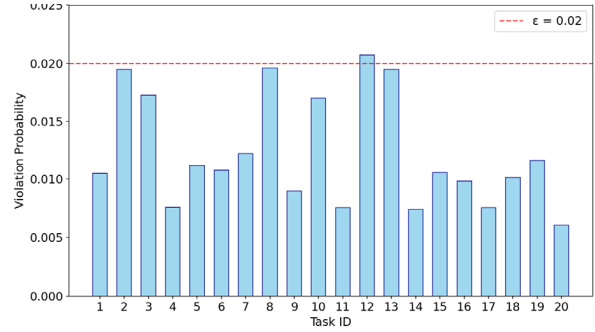

Figure 3: Violation probabilities for each task

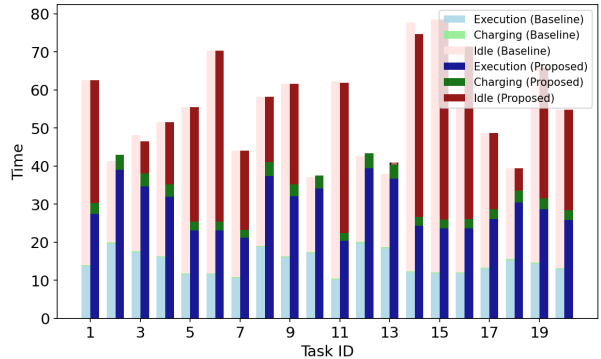

Figure 4: SAA time scheduling in 20 tasks testing

*4.2. Computational Time*

Most optimal control or nonlinear optimization problems, including the original formulation in this paper, are NP-hard and lack analytical solutions. To evaluate the computational efficiency of the proposed approach, a comparative analysis is conducted using different solution methods with 20 tasks. For the original nonlinear formulation, the solution using GA is established as the baseline. The GA serves as a reference point with a computational time of approximately 600 seconds to find a global solution. The linearized model with sample SAA implementation, which incorporates uncertainty handling, requires between 50 to 120 seconds depending on the number of samples used, while still maintaining solution quality within 0.5% of the GA.

## 5. SUMMARY AND CONCLUSIONS

This paper proposes a novel optimization framework to minimize battery degradation for AMRs. The degradation consists of not only charging transport aging but also the SOC during idle time. Tasks uncertainties are included in the optimization as well. The main contributions of this paper are: 1) Integrating battery calendar capacity loss or loss during idle time in AMRs planning optimization; 2) Establishing a first ever optimization framework that balances the objective of maintaining the lowest SOC and completing tasks as fast as possible; 3) Linearizing a nonlinear bilinear model with bilinear terms using piecewise McCormick Envelopes approach; and 4) Introducing a two-stage optimization model to handle the uncertainties and reformulating it into a tractable problem by appropriate approximation.

## REFERENCES

Ben-Tal, A., Goryashko, A., Guslitzer, E., and Nemirovski, A. (2004). Adjustable robust solutions of uncertain linear programs. *Mathematical Programming*, 99 (2), 351-376.

Castro, P. M. (2015). Tightening piecewise McCormick relaxations for bilinear problems. *Computers & Chemical Engineering*, 72, 300-311.

Choi, S. S. and Lim, H. S. (2002). Factors that affect cycle-life and possible degradation mechanisms of a Li-ion cell based on LiCoO2. *Journal of Power Sources*, 111 (1), 130-136.

Doyle, M., Fuller, T. F., and Newman, J. (1993). Modeling of galvanostatic charge and discharge of the lithium/polymer/insertion cell. *Journal of The Electrochemical Society*, 140 (6), 1526-1533.

Fuller, T. F., Doyle, M., and Newman, J. (1994). Simulation and optimization of the dual lithium ion insertion cell. *Journal of The Electrochemical Society*, 141 (1), 1-10.

Grolleau, S., et al. (2014). Calendar aging of commercial graphite/LiFePO4 cell - Predicting capacity fade under time dependent storage conditions. *Journal of Power Sources*, 255, 450-458.

Hoke, A., Brissette, A., Maksimović, D., Pratt, A. and Smith, K. (2011). Electric vehicle charge optimization including effects of lithium-ion battery degradation. In *2011 IEEE Vehicle Power and Propulsion Conference* (pp. 1-8). IEEE.

Kim, Yoon-Ho and Ha, Hoi-Doo (1997). Design of interface circuits with electrical battery models. *IEEE Transactions on Industrial Electronics*, 44 (1), 81-86.

Kivekas, K., Vepsalainen, J., and Tammi, K. (2018). Stochastic driving cycle synthesis for analyzing the energy consumption of a battery electric bus. *IEEE Access*, 6, 55586-55598.

Korolko, N. and Sahinoglu, Z. (2017). Robust optimization of EV charging schedules in unregulated electricity markets. *IEEE Transactions on Smart Grid*, 8 (1), 149-157.

Kuhn, D., Wiesemann, W., and Georghiou, A. (2009). Primal and dual linear decision rules in stochastic and robust optimization. *Mathematical Programming*, 130 (1), 177-209.

Lam, L. and Bauer, P. (2013). Practical capacity fading model for Li-ion battery cells in electric vehicles. *IEEE Transactions on Power Electronics*, 28 (12), 5910-5918.

Moss, P. L., Au, G., Plichta, E. J., and Zheng, J. P. (2008). An electrical circuit for modeling the dynamic response of Li-ion polymer batteries. *Journal of The Electrochemical Society*, 155 (12), A986.

Ning, G. and Popov, B. N. (2004). Cycle life modeling of lithium-ion batteries. *Journal of The Electrochemical Society*, 151 (10), A1584.

Ramadass, P., Haran, B., White, R., and Popov, B. N. (2003). Mathematical modeling of the capacity fade of Li-ion cells. *Journal of Power Sources*, 123 (2), 230-240.

Rong, P. and Pedram, M. (2006). An analytical model for predicting the remaining battery capacity of lithium-ion batteries. *IEEE Transactions on Very Large Scale Integration (VLSI) Systems*, 14 (5), 441-451.

Spotnitz, R. (2003). Simulation of capacity fade in lithium-ion batteries. *Journal of Power Sources*, 113 (1), 72-80.

Tredeau, F. P. and Salameh, Z. M. (2009). Evaluation of lithium iron phosphate batteries for electric vehicles application. In *2009 IEEE Vehicle Power and Propulsion Conference* (pp. 1266-1270). IEEE.